%% file: main.tex
\pdfoutput=1

\def\year{2018}\relax
\documentclass[letterpaper]{article} 
\usepackage{aaai18}  
\usepackage{times}  
\usepackage{helvet}  
\usepackage{courier}  
\usepackage{url}  
\usepackage{graphicx}  
\usepackage{amsmath}
\usepackage{amssymb}
\usepackage{mathtools}

\usepackage[linesnumbered,ruled]{algorithm2e}

\begin{document}

\title{OptionGAN: Learning Joint Reward-Policy Options using Generative Adversarial Inverse Reinforcement Learning}

\author{
Peter Henderson\textsuperscript{1},
Wei-Di Chang\textsuperscript{2},
Pierre-Luc Bacon\textsuperscript{1}\\
\Large{\bf{David Meger\textsuperscript{1},
Joelle Pineau\textsuperscript{1},
Doina Precup\textsuperscript{1}}}\\
\textsuperscript{1} School of Computer Science, McGill University, Montreal, Canada\\
\textsuperscript{2} Department of Electrical, Computer, and Software Engineering, McGill University, Montreal, Canada\\
\url{{peter.henderson,wei-di.chang}@mail.mcgill.ca}\\
\url{pbacon@cs.mcgill.ca},
\url{dmeger@cim.mcgill.ca},
\url{{jpineau,dprecup}@cs.mcgill.ca}
}

\maketitle

\begin{abstract}
Reinforcement learning has shown promise in learning policies that can solve complex problems. However, manually specifying a good reward function can be difficult, especially for intricate tasks. Inverse reinforcement learning offers a useful paradigm to learn the underlying reward function directly from expert demonstrations. Yet in reality, the corpus of demonstrations may contain trajectories arising from a diverse set of underlying reward functions rather than a single one. Thus, in inverse reinforcement learning, it is useful to consider such a decomposition. The options framework in reinforcement learning is specifically designed to decompose policies in a similar light. We therefore extend the options framework and propose a method to simultaneously recover reward options in addition to policy options. We leverage adversarial methods to learn joint reward-policy options using only observed expert states. We show that this approach works well in both simple and complex continuous control tasks and shows significant performance increases in one-shot transfer learning.
\end{abstract}

\section{Introduction}
A long term goal of Inverse Reinforcement Learning (IRL) is to be able to learn underlying reward functions and policies solely from human video demonstrations. We call such a case, where the demonstrations come from different contexts and the task must be performed in a novel environment, one-shot transfer learning. For example, given only demonstrations of a human walking on earth, can an agent learn to walk on the moon?

However, such demonstrations would undoubtedly come from a wide range of settings and environments and may not conform to a single reward function. This proves detrimental to current methods which might over-generalize and cause poor performance. In forward RL, decomposing a policy into smaller specialized policy options has been shown to improve results for exactly such cases~\cite{sutton1999between,bacon2016option}. Thus, we extend the options framework to IRL and decompose both the reward function and policy. Our method is able to learn deep policies which can specialize to the set of best-fitting experts. Hence, it excels at one-shot transfer learning where single-approximator methods waver.

To accomplish this, we make use of the Generative Adversarial Imitation Learning (GAIL) framework~\cite{GAIL} and formulate a method for learning joint reward-policy options with adversarial methods in IRL. As such, we call our method OptionGAN.
This method can implicitly learn divisions in the demonstration state space and accordingly learn policy and reward options.
Leveraging a correspondence between Mixture-of-Experts (MoE) and one-step options, we learn a decomposition of rewards and the policy-over-options in an end-to-end fashion.
This decomposition is able to capture simple problems and learn any of the underlying rewards in one shot. This gives flexibility and benefits for a variety of future applications (both in reinforcement learning and standard machine learning).

We evaluate OptionGAN in the context of continuous control locomotion tasks, considering both simulated MuJoCo locomotion OpenAI Gym environments~\cite{gym}, modifications of these environments for task transfer~\cite{gym-extensions}, and a more complex Roboschool task~\cite{PPO}. We show that the final policies learned using joint reward-policy options outperform a single reward approximator and policy network in most cases, and particularly excel at one-shot transfer learning.

\section{Related Work}
One goal in robotics research is to create a system which learns how to accomplish complex tasks simply from observing an expert's actions (such as videos of humans performing actions). While IRL has been instrumental in working towards this goal, it has become clear that fitting a single reward function which generalizes across many domains is difficult. To this end, several works investigate decomposing the underlying reward functions of expert demonstrations and environments in both IRL and RL~\cite{krishnan2016hirl,sermanet2016unsupervised,NIPS2012_4737,babes2011apprenticeship,van2017hybrid}. For example, in~\cite{krishnan2016hirl}, reward functions are decomposed into a set of subtasks based on segmenting expert demonstration transitions (known state-action pairs) by analyzing the changes in ``local linearity with respect to a kernel function''. Similarly, in~\cite{sermanet2016unsupervised}, techniques in video editing based on information-similarity are adopted to divide a video demonstration into distinct sections which can then be recombined into a differentiable reward function.

However, simply decomposing the reward function may not be enough, the policy must also be able to adapt to different tasks. Several works have investigated learning a latent dimension along with the policy for such a purpose~\cite{hausman2017multi,wang2017robust,li2017inferring}. This latent dimension allows multiple tasks to be learned by one policy and elicited via the latent variable. In contrast, our work focuses on one-shot transfer learning. In the former work, the desired latent variable must be known and provided, whereas in our formulation the latent structure is inherently encoded in an unsupervised manner. This is inherently accomplished while learning to solve a task composed of a wide range of underlying reward functions and policies in a single framework. Overall, this work contains parallels to all of the aforementioned and other works emphasizing hierarchical policies~\cite{daniel2012hierarchical,dietterich2000hierarchical,merel2017learning}, but specifically focuses on leveraging MoEs and reward decompositions to fit into the \emph{options} framework for efficient one-shot transfer learning in IRL.

\section{Preliminaries and Notation}

\textbf{Markov Decision Processes (MDPs)} MDPs consist of states $S$, actions $A$, a transition function $P : S \times A \rightarrow (S \rightarrow \mathbb{R}) $, and a reward function $r : S \rightarrow \mathbb{R}$. We formulate our methods in the space of continuous control tasks ($A \in \mathbb{R}, S \in \mathbb{R}$) using measure-theoretic assumptions. Thus we define a parameterized \textit{policy} as the probability distribution over actions conditioned on states $\pi_\theta : S \times A \rightarrow [0,1]$, modeled by a Gaussian $\pi_\theta \sim \mathcal{N}(\mu,\,\sigma^{2})$ where $\theta$ are the policy parameters. The value of a policy is defined as $V_\pi(s) = \mathbb{E}_\pi [ \sum_{t=0}^\infty \gamma^t r_{t+1} | s_0 = s]$ and the action-value is $Q_\pi(s,a) = \mathbb{E}_\pi [\sum_{t=0}^\infty \gamma^t r_{t+1} | s_0 = s, a_0 = a]$, where $\gamma \in [0,1)$ is the discount factor.

\textbf{The Options framework} In reinforcement learning, an option ($\omega \in \Omega$) can be defined by a triplet ($I_\omega, \pi_\omega, \beta_\omega$). In this definition, $\pi_\omega$ is called an intra-policy option, $I_\omega \subseteq S$ is an initiation set, and $\beta_\omega : S \rightarrow [0,1]$ is a termination function (i.e. the probability that an option ends at a given state)~\cite{sutton1999between}. Furthermore, $\pi_\Omega$ is the policy-over-options. That is, $\pi_\Omega$ determines which option $\pi_\omega$ an agent picks to use until the termination function $\beta_\omega$ indicates that a new option should be chosen. Other works explicitly formulate \textit{call-and-return} options, but we instead simplify to \textit{one-step} options, where $\beta_\omega (s) = 1; \forall \omega \in \Omega, \forall s \in S$. One-step options have long been discussed as an alternative to temporally extended methods and often provide advantages in terms of optimality and value estimation~\cite{sutton1999between,dietterich2000hierarchical,daniel2012hierarchical}.
Furthermore, we find that our options still converge to temporally extended and interpretable actions.

\textbf{Mixture-of-Experts} The idea of creating a mixture of experts (MoEs) was initially formalized to improve learning of neural networks by dividing the input space among several networks and then combining their outputs through a soft weighted average~\cite{expertmixtures-91}. It has since come into prevalence for generating extremely large neural networks~\cite{superlargenets-17}. In our formulation of joint reward-policy options, we leverage a correspondence between Mixture-of-Experts and options. In the case of one-step options, the policy-over-options ($\pi_\Omega$) can be viewed as a specialized gating function over experts (intra-options policies $\pi_\omega(a|s)$): $\sum_\omega \pi_\Omega (\omega| s) \pi_\omega(a | s)$. Several works investigate convergence to a sparse and specialized Mixture-of-Experts~\cite{expertmixtures-91,superlargenets-17}. We leverage these works to formulate a Mixture-of-Experts which converges to one-step options.


\textbf{Policy Gradients} Policy gradient (PG) methods \cite{sutton2000policy} formulate a method for optimizing a parameterized policy $\pi_\theta$ through stochastic gradient ascent. In the discounted setting, PG methods optimize  $\rho (\theta, s_0) = \mathbb{E}_{\pi_\theta}\left[ \sum_{t=0}^\infty \gamma^t r(s_t) | s_0 \right]$.
The PG theorem states: $\frac{\delta \rho (\theta, s_0)}{\delta \theta} = \sum_s \mu_{\pi_\theta} (s | s_0) \sum_a \frac{\delta \pi_\theta (a | s)}{\delta \theta} Q_{\pi_\theta} (s,a)$, where $\mu_{\pi_\theta} (s | s_0) = \sum_{t=0}^\infty \gamma^t P(s_t = s | s_0)$ \cite{sutton2000policy}. In Trust Region Policy Optimization (TRPO) \cite{TRPO} and Proximal Policy Optimization (PPO) \cite{PPO} this update is constrained and transformed into the advantage estimation view such that the above becomes a constrained optimization: $\max_\theta \mathbb{E}_{t} \left[ \frac{\pi_\theta (a_t|s_t)}{\pi_{\theta_{old}}(a_t|s_t)} A_t (s_t, a_t) \right]$ subject to $\mathbb{E}_t \left[\text{KL} \left[ \pi_{\theta_{old}} ( \cdot | s_t), \pi_\theta (\cdot | s_t) \right] \right] \le \delta$ where $A_t (s_t, a_t)$ is the generalized advantage function according to \cite{schulman2015high}.
In TRPO, this is solved as a constrained conjugate gradient descent problem, while in PPO the constraint is transformed into a penalty term or clipping objective.

\begin{figure*}[!htb]
\centering
\includegraphics[angle=90, width=0.4\textwidth]{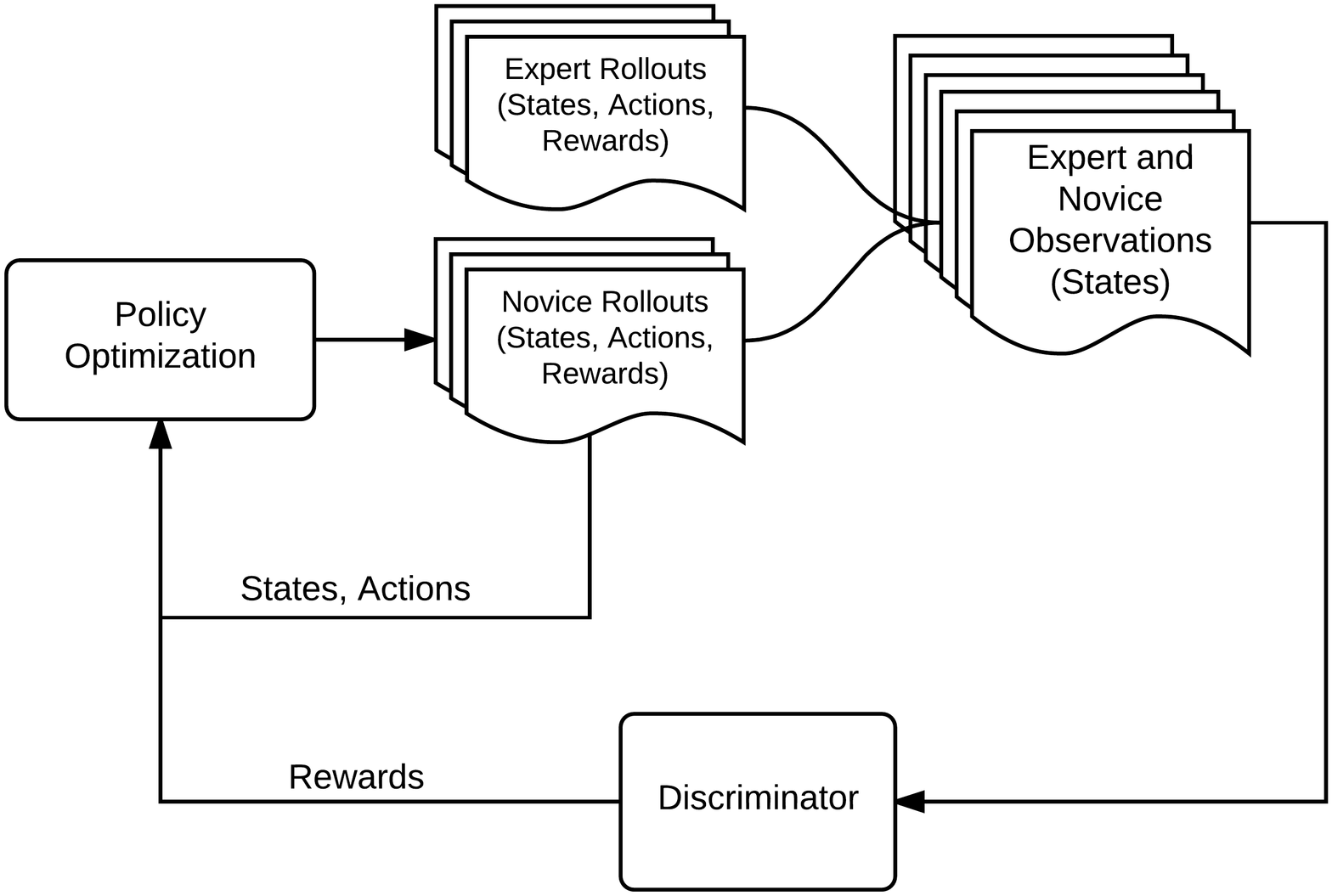}
\includegraphics[angle=90, width=0.4\textwidth]{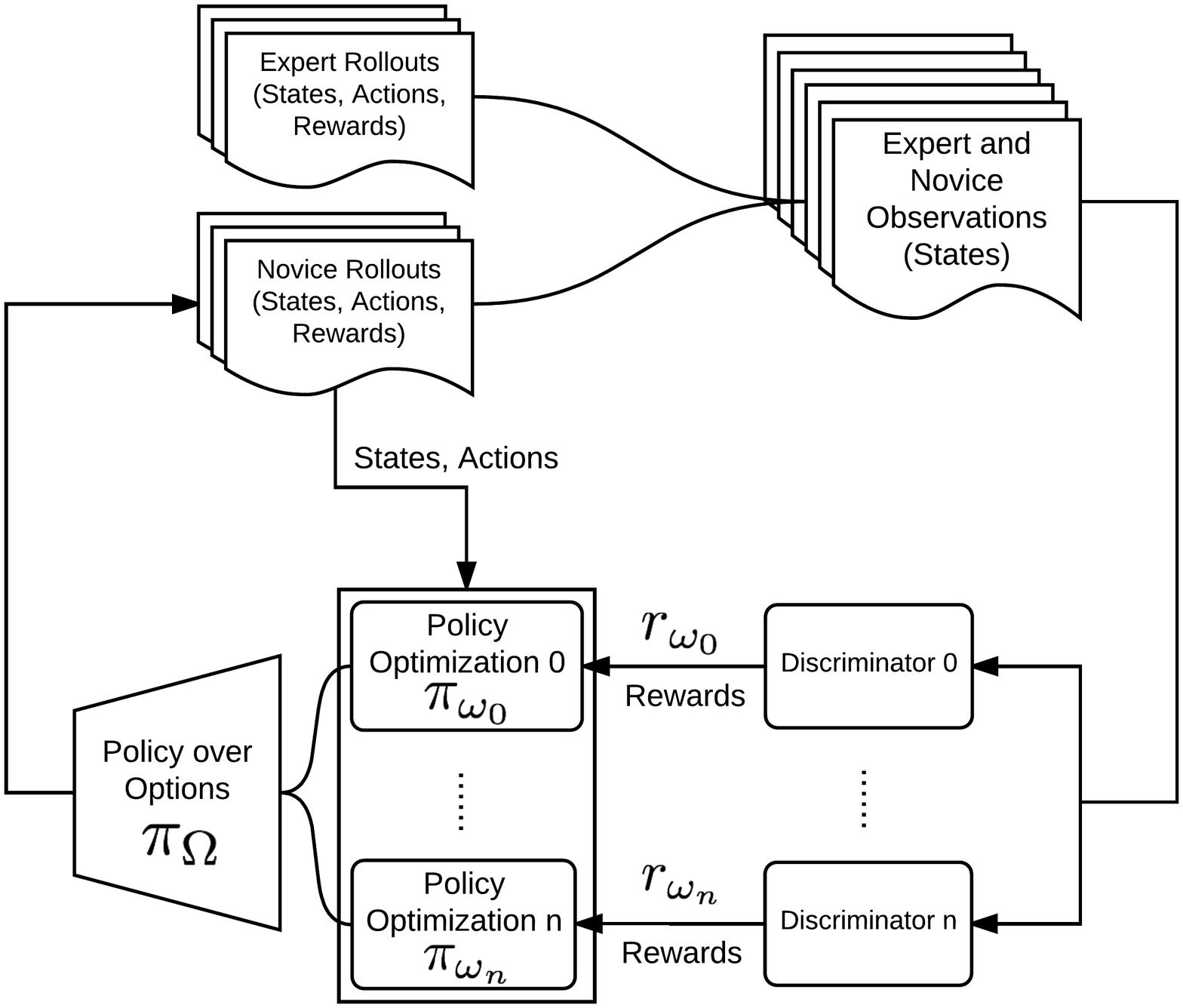}
\caption{Generative Adversarial Inverse Reinforcement Learning (left) and OptionGAN (right) Architectures}
\label{fig:architecture}
\end{figure*}

\textbf{Inverse Reinforcement Learning} Inverse Reinforcement Learning was first formulated in the context of MDPs by \cite{ng2000}. In later work, a parametrization of the reward function is learned as a linear combination of the state feature expectation so that the hyperdistance between the expert and the novice's feature expectation is minimized~\cite{apprenticeship2004}. It has also been shown that a solution can be formulated using the maximum entropy principle, with the goal of matching feature expectation as well~\cite{ziebart-maxent}.
Generative adversarial imitation learning (GAIL) make use of adversarial techniques from~\cite{goodfellow2014generative} to perform a similar feature expectation matching~\cite{GAIL}. In this case, a discriminator uses state-action pairs (transitions) from the expert demonstrations and novice rollouts to learn a binary classification probability distribution. The probability that a state belongs to an expert demonstration can then be used as the reward for a policy optimization step. However, unlike GAIL, we do not assume knowledge of the expert actions. Rather, we rely solely on observations in the discriminator problem. We therefore refer to our baseline approach as Generative Adversarial Inverse Reinforcement Learning (IRLGAN) as opposed to imitation learning. It is important to note that IRLGAN is GAIL without known actions, we adopt the different naming scheme to highlight this difference. As such, our adversarial game optimizes:
\begin{equation}
\max_{\pi_\Theta}\min_{R_{\hat{\Theta}}} - \left[\mathbb{E}_{\pi_\Theta} [ \log R_{\hat{\Theta}} (s)] + \mathbb{E}_{\pi_E} [\log (1- R_{\hat{\Theta}} (s))] \right]
\end{equation}

\noindent where $\pi_\Theta$ and $\pi_E$ are the policy of the novice and expert parameterized by $\Theta$ and $E$, respectively, and $R_{\hat{\Theta}}$ is the discriminator probability that a sample state belongs to an expert demonstration (parameterized by ${\hat{\Theta}}$). We use this notation since in this case the discriminator approximates a reward function. Similarly to GAIL, we use TRPO during the policy optimization step for simple tasks. However, for complex tasks we adopt PPO. Figure~\ref{fig:architecture} and Algorithm~\ref{algo:irlgan} show an outline for the general IRLGAN process.

\begin{algorithm}
    \SetKwInOut{Input}{Input}
    \SetKwInOut{Output}{Output}

    \Input{Expert trajectories $\tau_E \sim \pi_E$.}
    Initialize $\Theta, {\hat{\Theta}}$ \\
    \For{$i=0,1,2,\dots$}
    {
    Sample trajectories $\tau_N \sim \pi_{\Theta_i}$\\
    Update discriminator parameters (${\hat{\Theta}}$) according to:
    $$L_{\hat{\Theta}} = \mathbb{E}_{s \sim \tau_N} [ \log R_{\hat{\Theta}} (s) ] +\mathbb{E}_{s \sim \tau_E} [ \log(1 - R_{\hat{\Theta}} (s)) ] $$

    Update policy (with constrained update step and parameters $\theta$) according to:
    $$\mathbb{E}_{\tau_N} [ \nabla_\Theta \log \pi_{\Theta_i} (a | s) \mathbb{E}_{\tau_N} [ \log ( R_{{\hat{\Theta}}_{i+1}} (s)) | s_0 = \bar{s}]]$$}
    \caption{IRLGAN}
    \label{algo:irlgan}
\end{algorithm}

\section{Reward-Policy Options Framework}

Based on the need to infer a decomposition of underlying reward functions from a wide range of expert demonstrations in one-shot transfer learning, we extend the options framework for decomposing rewards as well as policies. In this way, intra-option policies, decomposed rewards, and the policy-over-options can all be learned in concert in a cohesive framework. In this case, an option is formulated by a tuple: ($I_\omega, \pi_\omega, \beta_\omega, r_\omega$). Here, $r_\omega$ is a reward option from which a corresponding intra-option policy $\pi_\omega$ is derived.
That is, each policy option is optimized with respect to its own local reward option. The policy-over-options not only chooses the intra-option policy, but the reward option as well: $\pi_\Omega \rightarrow (r_\omega, \pi_\omega)$. For simplicity, we refer to the policy-over-reward-options as $r_\Omega$ (in our formulation, $r_\Omega = \pi_\Omega$).
There is a parallel to be drawn from this framework to Feudal RL~\cite{dayan1993feudal}, but here the intrinsic reward function is statically bound to each worker (policy option), whereas in that framework the worker dynamically receives a new intrinsic reward from the manager.

To learn joint reward-policy options, we present a method which fits into the framework of IRLGAN. We reformulate the discriminator as a Mixture-Of-Experts and re-use the gating function when learning a set of policy options. We show that by properly formulating the discriminator loss function, the Mixture-Of-Experts converges to one-step options. This formulation also allows us to use regularizers which encourage distribution of information, diversity, and sparsity in both the reward and policy options.

\section{Learning Joint Reward-Policy Options}

The use of one-step options allows us to learn a policy-over-options in an end-to-end fashion as a Mixture-of-Experts formulation. In the one-step case, selecting an option ($\pi_{\omega,\theta}$) using the policy-over-options ($\pi_{\Omega,\zeta}$) can be viewed as a mixture of completely specialized experts such that: $\pi_\Theta(a|s) = \sum_\omega \pi_{\Omega,\zeta}(\omega| s) \pi_{\omega,\theta}(a|s)$.
The reward for a given state is composed as: $R_{\Omega, {\hat{\Theta}}}(s) = \sum_\omega \pi_{\Omega,\zeta} (\omega| s) r_{\omega, {\hat{\theta}}}(s)$,
where $\zeta, \theta \in \Theta, {\hat{\theta}} \in {\hat{\Theta}}$ are the parameters of the policy-over-options, policy options, and reward options, respectively. Thus, we reformulate our discriminator loss as a weighted mixture of completely specialized experts in Eq.~\ref{eq:discriminator_loss}. This allows us to update the parameters of the policy-over-options and reward options together during the discriminator update.

\begin{equation}
\begin{split}
L_\Omega &= \mathbb{E}_\omega \left[ \pi_{\Omega,\zeta}(\omega|s) L_{{\hat{\theta}},\omega} \right] + L_{reg}
\end{split}
\label{eq:discriminator_loss}
\end{equation}

Here, $L_{{\hat{\theta}},\omega}$ is the sigmoid cross-entropy loss of the reward options (discriminators). $L_{reg}$, as will be discussed later on, is a penalty or set of penalties which can encourage certain properties of the policy-over-options or the overall reward signal. As can be seen in Algorithm~\ref{algo:OptionGAN} and Figure~\ref{fig:architecture}, this loss function can fit directly into the IRLGAN framework.

\begin{algorithm}
    \SetKwInOut{Input}{Input}
    \SetKwInOut{Output}{Output}

    \Input{Expert trajectories $\tau_E \sim \pi_E$.}
    Initialize $\theta, {\hat{\theta}}$ \\
    \For{$i=0,1,2,\dots$}
    {
    Sample trajectories $\tau_N \sim \pi_{\Theta_i}$\\
    Update discriminator options parameters ${\hat{\theta}},\omega$ and policy-over-options parameters $\zeta$, to minimize:
    $$L_\Omega = \mathbb{E}_\omega \left[ \pi_{\Omega,\zeta}(\omega|s) L_{{\hat{\theta}},\omega} \right] + L_{reg} $$
    $$L_{{\hat{\theta}},\omega} = \mathbb{E}_{\tau_N} [ \log r_{{\hat{\theta}},\omega} (s) ] +\mathbb{E}_{\tau_E} [ \log(1 - r_{{\hat{\theta}},\omega} (s)) ]$$

    Update policy options (with constrained update step and parameters $\theta_\omega \in \Theta_\Omega$) according to:
    $$\mathbb{E}_{\tau_N} [ \nabla_\theta \log \pi_\Theta (a | s) \mathbb{E}_{\tau_N} [ \log ( R_{\Omega, {\hat{\Theta}}} (s)) | s_0 = \bar{s}]]$$}
    \caption{OptionGAN}
    \label{algo:OptionGAN}
\end{algorithm}

Having updated the parameters of the policy-over-options and reward options, standard PG methods can be used to optimize the parameters of the intra-option policies. This can be done by weighting the average of the intra-option policy actions with the policy-over-options $\pi_{\Omega,\zeta}$.
While it is possible to update each intra-option policy separately as in~\cite{bacon2016option}, this Mixture-of-Experts formulation is equivalent, as discussed in the next section. Once the gating function specializes over the options, all gradients except for those related to the intra-option policy selected would be weighted by zero. We find that this end-to-end parameter update formulation leads to easier implementation and smoother learning with constraint-based methods.

\section{Mixture-of-Experts as Options}

To ensure that our MoE formulation converges to options in the optimal case, we must properly formulate our loss function such that the gating function specializes over experts. While it may be possible to force a sparse selection of options through a top-$k$ choice as in~\cite{superlargenets-17}, we find that this leads to instability since for $k=1$ the top-$k$ function is not differentiable. As is specified in~\cite{expertmixtures-91}, a loss function of the form $L = (y - \frac{1}{||\Omega||} \sum_\omega \pi_\Omega(\omega| s) y_\omega(s))^2$ draws cooperation between experts, but a reformulation of the loss, $L = \frac{1}{||\Omega||} \sum_\omega \pi_\Omega(\omega| s)(y - y_\omega(s))^2$, encourages specialization.


If we view our policy-over-options as a softmax (i.e. $\pi_\Omega(\omega| s) = \frac{\exp(z_\omega(s))}{\sum_i \exp(z_i(s))}$), then the derivative of the loss function with respect to the gating function becomes:

\begin{equation}
    \frac{dL}{dz_\omega} = \frac{1}{||\Omega||} \pi_\Omega(\omega| s) \left( (y - y_\omega(s))^2 - L \right)
\end{equation}

This can intuitively be interpreted as encouraging the gating function to increase the likelihood of choosing an expert when its loss is less than the average loss of all the experts. The gating function will thus move toward deterministic selection of experts.

As we can see in Eq.~\ref{eq:discriminator_loss}, we formulate our discriminator loss in the same way, using each reward option and the policy-over-options as the experts and gating function respectively. This ensures that the policy-over-options specializes over the state space and converges to a deterministic selection of experts. Hence, we can assume that in the optimal case, our formulation of an MoE-style policy-over-options is equivalent to one-step options. Our characterization of this notion of MoE-as-options is further backed by experimental results. Empirically, we still find temporal coherence across option activation despite not explicitly formulating call-and-return options as in~\cite{bacon2016option}.

\subsection{Regularization Penalties}

\begin{figure*}[!ht]
\centering\hspace{-.2cm}\includegraphics[width=.96\textwidth]{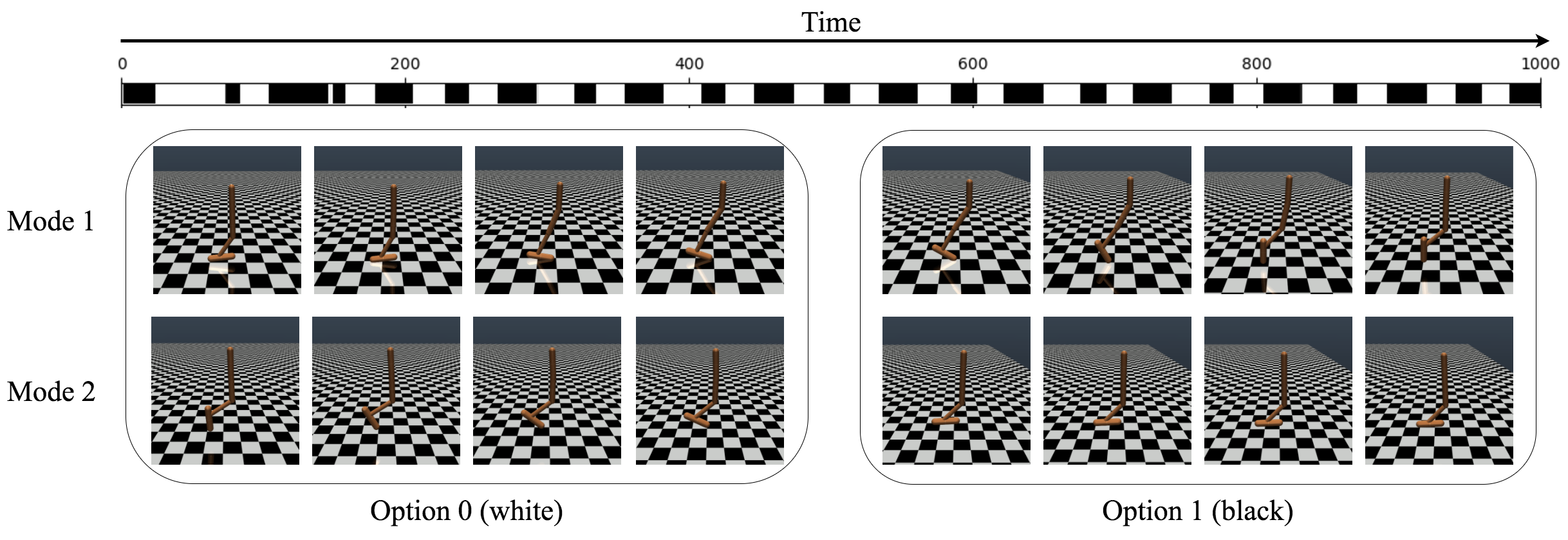}
\caption{The policy-over-options elicits two interpretable behaviour modes per option, but temporal cohesion and specialization is seen between these behaviour modes across time within a sample rollout trajectory.}
\label{fig:optionexplained}
\end{figure*}

Due to our formulation of Mixture-of-Experts as options, we can learn our policy-over-options in an end-to-end manner. This allows us to add additional terms to our loss function to encourage the appearance of certain target properties.

\subsubsection{Sparsity and Variance Regularization    }    To ensure an even distribution of activation across the options, we look to conditional computation techniques that encourage sparsity and diversity in hidden layer activations and apply these to our policy-over-options~\cite{bengio2015conditional}. We borrow three penalty terms $L_b$, $L_e$, $L_v$ (adopting a similar notation). In the minibatch setting, these are formulated as:

\begin{eqnarray}
    L_b &=& \sum_\omega || \mathbb{E}_s[\pi_\Omega(\omega | s)] - \tau ||_2\\
    L_e &=& \mathbb{E}_s\left[ ||  \left(\frac{1}{||\Omega||} \sum_\omega \pi_\Omega(\omega | s) \right) - \tau ||_2 \right]\\
    L_v &=& - \sum_\omega var_\omega\{\pi_\Omega(\omega|s)\}
\end{eqnarray}

\noindent where $\tau$ is the target sparsity rate (which we set to $.5$ for all cases). Here, $L_b$ encourages the activation of the policy-over-options with target sparsity $\tau$ ``in expectation over the data''~\cite{bengio2015conditional}. Essentially, $L_b$ encourages a uniform distribution of options over the data while $L_e$ drives toward a target sparsity of activations per example (doubly encouraging our mixtures to be sparse). $L_v$ also encourages varied $\pi_\Omega$ activations while discouraging uniform selection.

\subsubsection{Mutual Information Penalty   }    To ensure the specialization of each option to a specific partition of the state space, a mutual information (MI) penalty is added.\footnote{While it may be simpler to use an entropy regularizer, we found that in practice it performs worse. Entropy regularization encourages exploration~\cite{mnih2016asynchronous}. In the OptionGAN setting, this results in unstable learning, while the mutual information term encourages diversity in the options while providing stable learning.} We thus minimize mutual information pairwise between option distributions, similarly to~\cite{liu2002learning}:
\begin{equation}
I(F_i; F_j) = -\frac{1}{2} \log (1-\rho_{ij}^2),
\end{equation}
where $F_i$ and $F_j$ are the outputs of reward options $i$ and $j$ respectively, and $\rho_{ij}$ the correlation coefficient of $F_i$ and $F_j$, defined as $\rho_{ij} = \frac{E[(F_i - E[F_i])(F_j - E[F_j])]}{\sigma_i^2 \sigma_j^2}$.

\noindent The resulting loss term is thus computed as:
\begin{equation}
    L_{\text{MI}} = \sum_{\omega \in \Omega} \sum_{\hat{\omega} \in \Omega, \omega\neq \hat{\omega}} I(\pi_\omega, \pi_{\hat{\omega}}).\
\end{equation}

\noindent Thus the overall regularization term becomes:
\begin{equation}
\mathcal{L}_{reg} = \lambda_b L_b + \lambda_e L_e + \lambda_v L_v + \lambda_{\text{MI}} L_{\text{MI}}.
\end{equation}

\section{Experiments}

\begin{table*}[!htb]
    \centering{\footnotesize
    \begin{tabular}{| c | c | c c c |}
    \hline
         Task & Expert & IRLGAN & OptionGAN (2ops) & OptionGAN (4ops) \\
         \hline
         Hopper-v1 & 3778.8 $\pm$ 0.3 &\textbf{3736.3 $\pm$ 152.4} &\textit{3641.2 $\pm$ 105.9}&\textit{3715.5 $\pm$ 17.6}\\
         HalfCheetah-v1 & 4156.9 $\pm$ 8.7 & 3212.9 $\pm$ 69.9& \textbf{3714.7 $\pm$ 87.5}& \textit{3616.1 $\pm$ 127.3}\\
         Walker2d-v1 & 5528.5 $\pm$ 7.3 & \textit{4158.7 $\pm$ 247.3}& \textit{3858.5 $\pm$ 504.9} & \textbf{4239.3 $\pm$ 314.2}\\
         \hline
         Hopper (One-Shot) & 3657.7 $\pm$ 25.4 & 2775.1 $\pm$ 203.3 & \textit{3409.4 $\pm$ 80.8} & \textbf{3464.0 $\pm$ 67.8}\\
         HalfCheetah (One-Shot) &4156.9 $\pm$ 51.3 & 1296.3 $\pm$ 177.8 & 1679.0 $\pm$ 284.2&\textbf{2219.4 $\pm$ 231.8}\\
         Walker (One-Shot) &4218.1 $\pm$ 43.1 & 3229.8 $\pm$ 145.3&\textbf{3925.3 $\pm$ 138.9} & \textit{3769.40 $\pm$ 170.4}\\
         \hline
         HopperSimpleWall-v0 & 3218.2 $\pm$ 315.7 & \textit{2897.5 $\pm$ 753.5} & \textit{3140.3 $\pm$ 674.3} & \textbf{3272.3 $\pm$ 569.0}\\
         RoboschoolHumanoidFlagrun-v1 &2822.1 $\pm$ 531.1 & \textit{1455.2 $\pm$ 567.6} & \textit{1868.9 $\pm$ 723.7} & \textbf{2113.6 $\pm$ 862.9}\\
         \hline
    \end{tabular}}
    \caption{True Average Return with the standard error across 10 trials on the 25 final evaluation rollouts using the final policy.}
    \label{tab:overall_results}
\end{table*}

To evaluate our method of learning joint reward-policy options, we investigate continuous control tasks. We divide our experiments into 3 settings: simple locomotion tasks, one-shot transfer learning, and complex tasks.
We compare OptionGAN against IRLGAN in all scenarios, investigating whether dividing the reward and policy into options improves performance against the single approximator case.\footnote{Extended experimental details and results can be found in the supplemental. Code is located at: \\ \url{https://github.com/Breakend/OptionGAN}.} Table~\ref{tab:overall_results} shows the overall results of our evaluations and we highlight a subset of learning curves in Figure~\ref{fig:gravtransfer}. We find that in nearly every setting, the final optionated policy learned by OptionGAN outperforms the single approximator case.


\subsection{Experimental Setup}
All shared hyperparameters are held constant between IRLGAN and OptionGAN evaluation runs. All evaluations are averaged across 10 trials, each using a different random seed. We use the average return of the true reward function across 25 sample rollouts as the evaluation metric. Multilayer perceptrons are used for all approximators as in~\cite{GAIL}. For the OptionGAN intra-option policy and reward networks, we use shared hidden layers. That is $r_\omega, \forall \omega \in \Omega$ all share hidden layers and $\pi_\omega, \forall \omega \in \Omega$ share hidden layers. We use separate parameters for the policy-over-options $\pi_\Omega$.
Shared layers are used to ensure a fair comparison against a single network of the same number of hidden layers.
For simple settings all hidden layers are of size $(64,64)$ and for complex experiments are $(128,128)$.
For the 2-options case we set $\lambda_e=10.0, \lambda_b=10.0, \lambda_v=1.0$ based on a simple hyperparameter search and reported results from~\cite{bengio2015conditional}.
For the 4-options case we relax the regularizer that encourages a uniform distribution of options ($L_b$), setting $\lambda_b=.01$.

\subsection{Simple Tasks}

First, we investigate simple settings without transfer learning for a set of benchmark locomotion tasks provided in OpenAI Gym \cite{gym} using the MuJoCo simulator \cite{mujoco}. We use the Hopper-v1, HalfCheetah-v1, and Walker2d-v1 locomotion environments. The results of this experiment are shown in Table~\ref{tab:overall_results} and sample learning curves for Hopper and HalfCheetah can be found in Figure~\ref{fig:gravtransfer}. We use 10 expert rollouts from a policy trained using TRPO for 500 iterations.

In these simple settings, OptionGAN converges to policies which perform as well or better than the single approximator setting. Importantly, even in these simple settings, the options which our policy selects have a notion of temporal coherence and interpretability despite not explicitly enforcing this in the form of a termination function. This can be seen in the two option version of the Hopper-v1 task in Figure~\ref{fig:optionexplained}. We find that generally each option takes on two behaviour modes. The first option handles: (1) the rolling of the foot during hopper landing; (2) the folding in of the foot in preparation for floating. The second option handles: (1) the last part of take-off where the foot is hyper-extended and body flexed; (2) the part of air travel without any movement.

\subsection{One-Shot Transfer Learning}

We also investigate one-shot transfer learning. In this scenario, the novice is trained on a target environment, while expert demonstrations come from a similar task, but from environments with altered dynamics (i.e. one-shot transfer from varied expert demonstrations to a new environment). To demonstrate the effectiveness of OptionGAN in these settings, we use expert demonstrations from environments with varying gravity conditions as seen in \cite{gym-extensions,christiano2016transfer}.
We vary the gravity (.5, .75, 1.25, 1.5 of Earth's gravity) and train experts using TRPO for each of these.
We gather 10 expert trajectories from each gravity variation, for a total of 40 expert rollouts, to train a novice agent on the normal Earth gravity environment (the default -v1 environment as provided in OpenAI Gym). We repeat this for Hopper-v1, HalfCheetah-v1, and Walker2D-v1.

These gravity tasks are selected due to the demonstration in \cite{gym-extensions} that learning sequentially on these varied gravity environments causes catastrophic forgetting of the policy on environments seen earlier in training.
This suggests that the dynamics are varied enough that trajectories are difficult to generalize across, yet still share some state representations and task goals. As seen in Figure~\ref{fig:gravtransfer}, using options can cause significant performance increases in this area, but performance gains can vary across the number of options and the regularization penalty as seen in Table~\ref{tab:overall_results}.

\subsection{Complex Tasks}

Lastly, we investigate slightly more complex tasks. We utilize the HopperSimpleWall-v0 environment provided by the gym-extensions framework \cite{gym-extensions} and the RoboschoolHumanoidFlagrun-v1 environment used in \cite{PPO}.
In the first, a wall is placed randomly in the path of the Hopper-v1 agent and simplified sensor readouts are added to the observations as in \cite{wang2017robust}.
In the latter, the goal is to run and reach a frequently changing target.
This is an especially complex task with a highly varied state space. In both cases we use an expert trained with TRPO and PPO respectively, to generate 40 expert rollouts.
For the Roboschool environment, we find that TRPO does not allow enough exploration to perform adequately, and thus we switch our policy optimization method to the clipping-objective version of PPO.

\section{Ablation Investigations}
\begin{figure*}
\includegraphics[width=.33\textwidth]{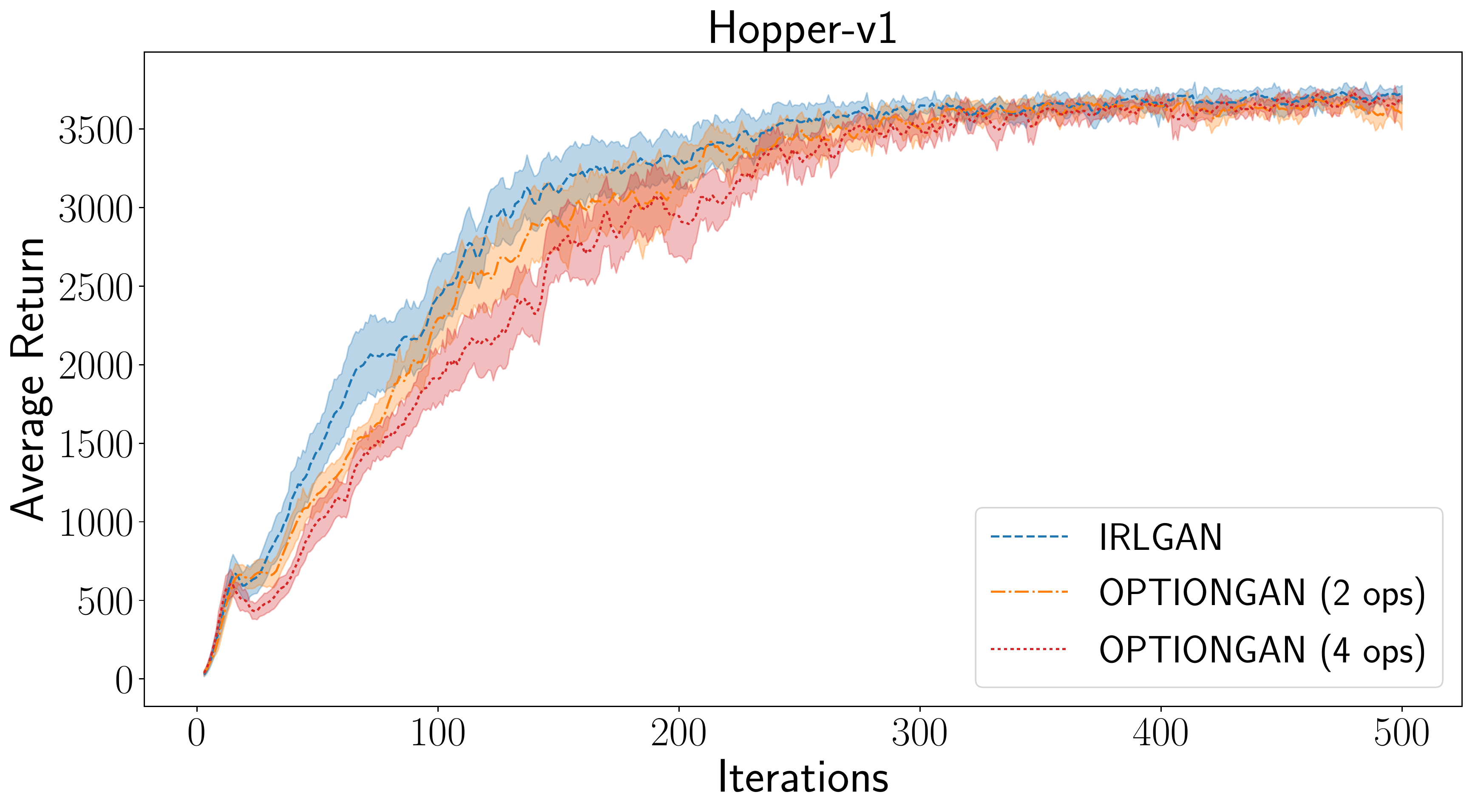}
\includegraphics[width=.33\textwidth]{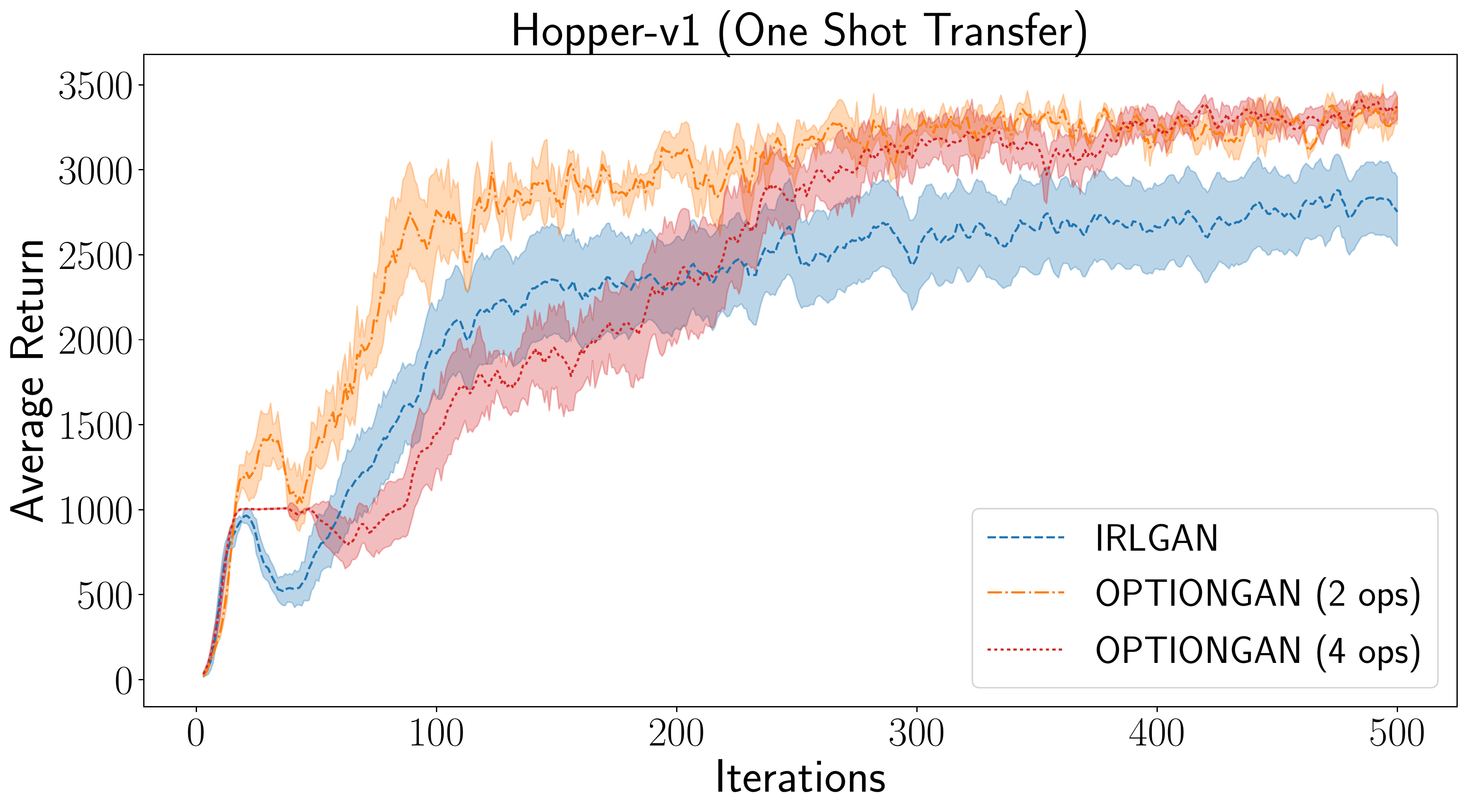}
\includegraphics[width=.33\textwidth]{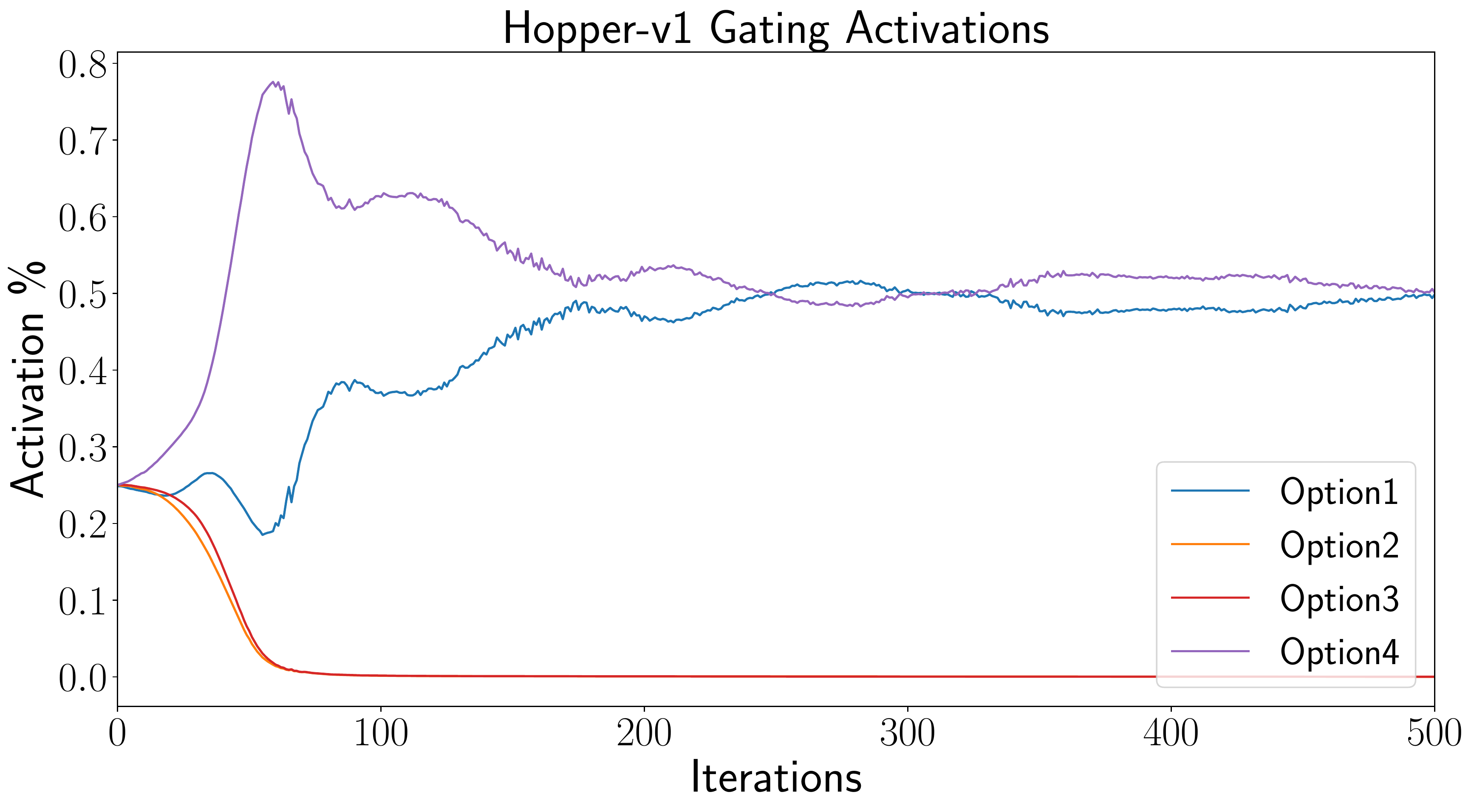}
\includegraphics[width=.33\textwidth]{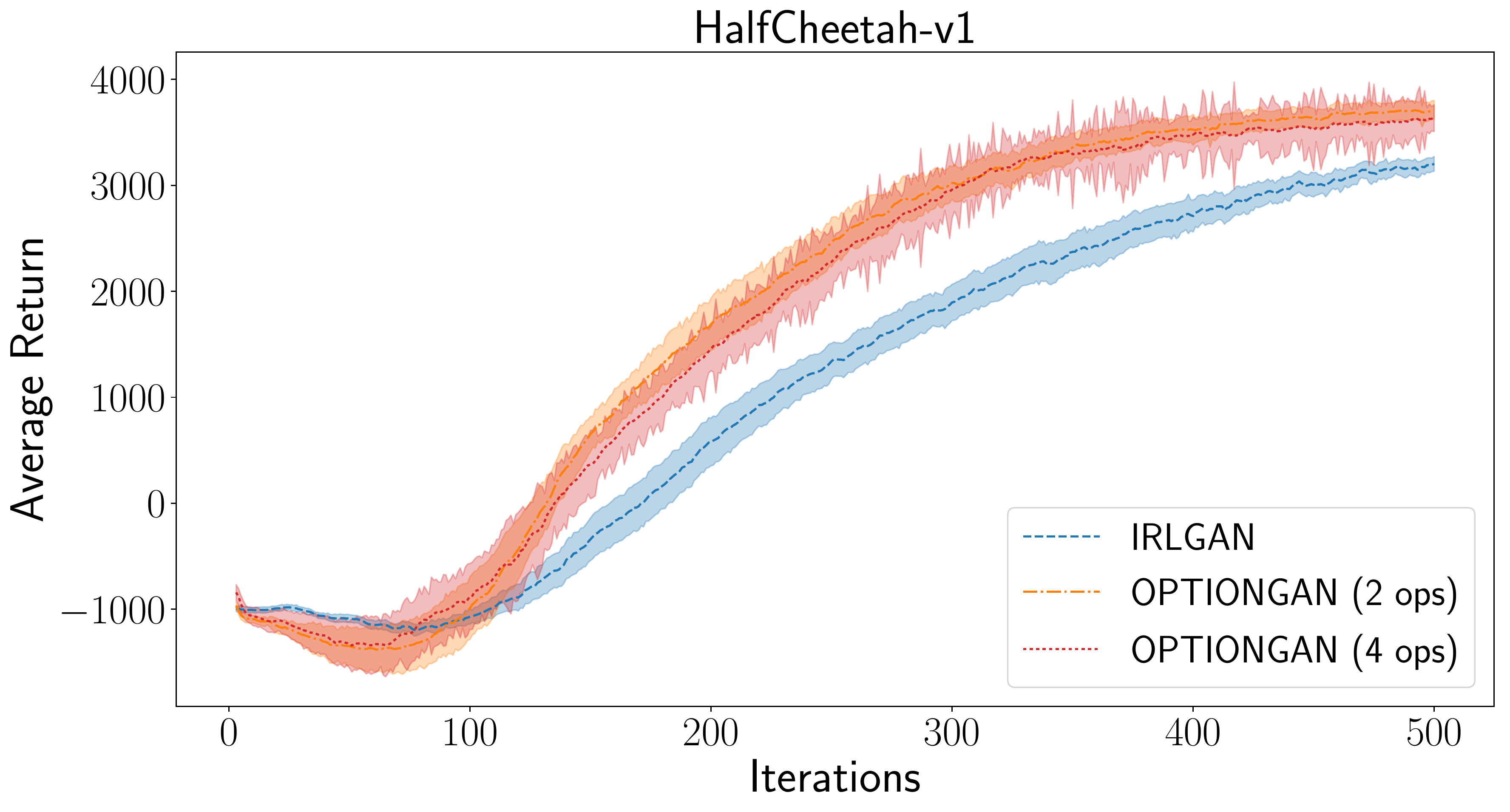}
\includegraphics[width=.33\textwidth]{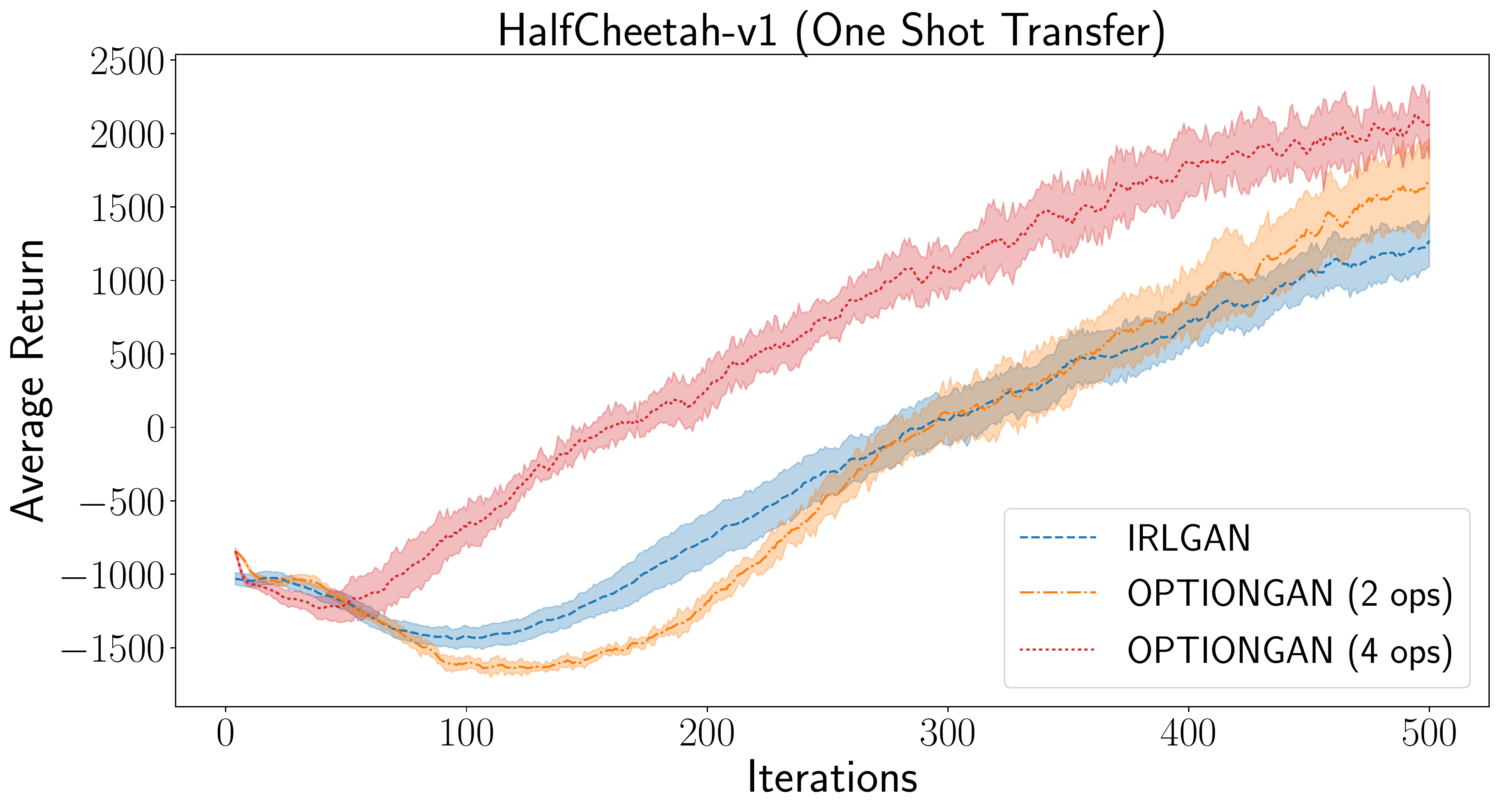}
\includegraphics[width=.33\textwidth]{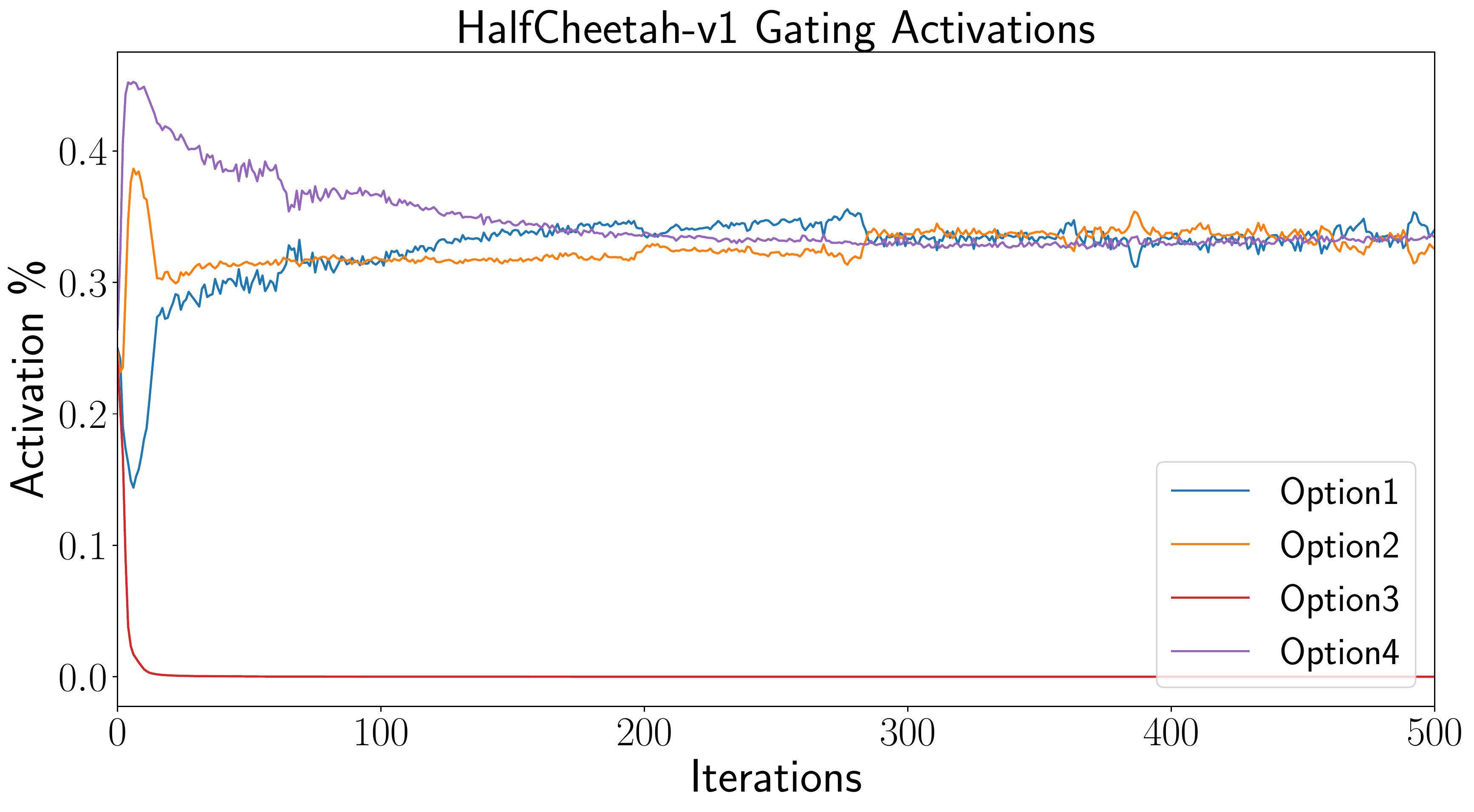}
\caption{ Left Column: Simple locomotion curves. Error bars indicate standard error of average returns across 10 trials on 25 evaluation rollouts. Middle Column: One-shot transfer experiments with 40 expert demonstrations from varied gravity environments without any demonstrations on the novice training environment training on demonstrations from .5G, .75G, 1.25G, 1.5G gravity variations. Right Column: Activations of policy-over-options over time with 4 options on training samples in the one-shot transfer setting with $\lambda_b=.01$.}
\label{fig:gravtransfer}
\end{figure*}

\begin{figure}
    \centering\includegraphics[width=.43\textwidth]{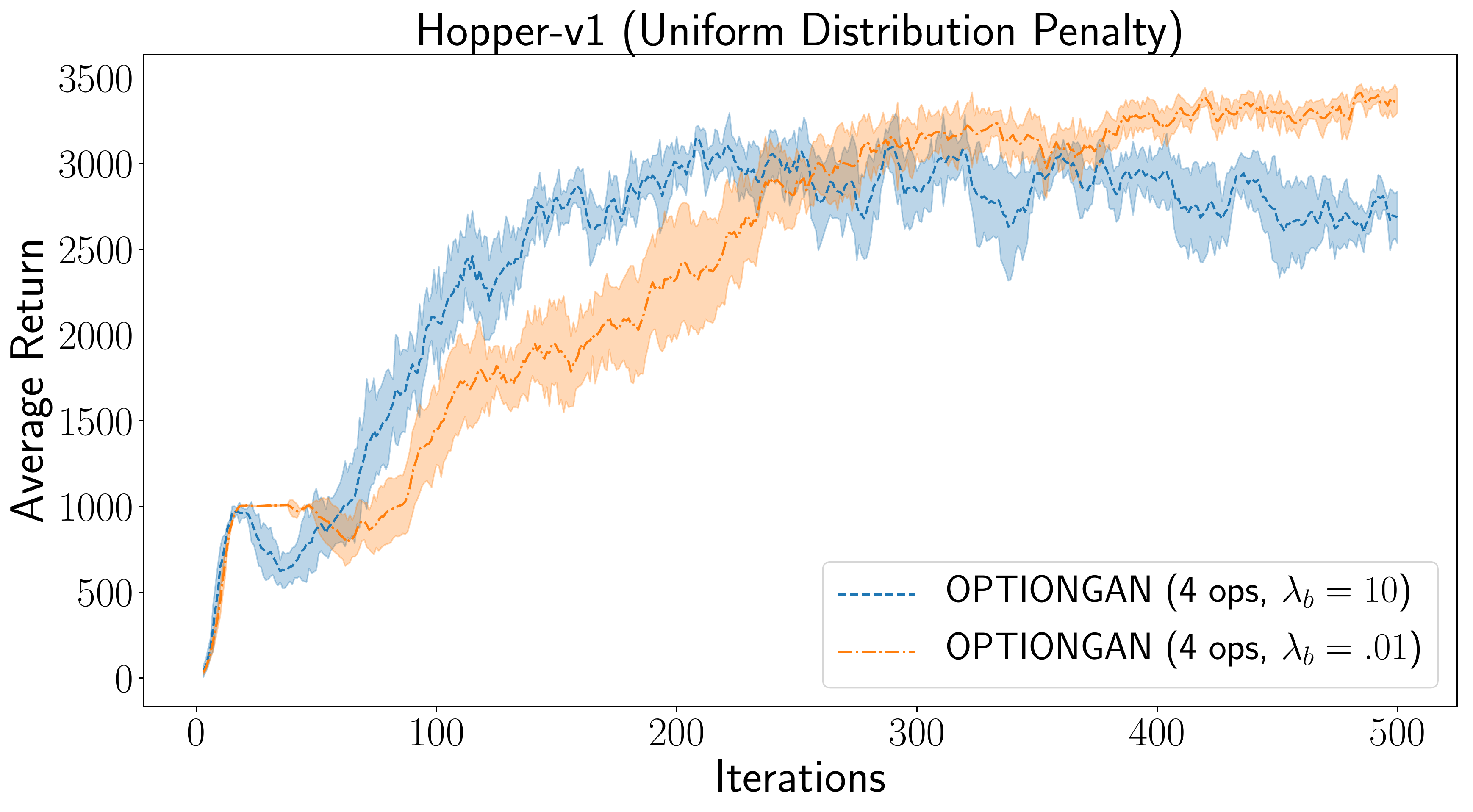}
    \caption{Effect of uniform distribution regularizer. Average $\pi_\Omega$ across final sample novice rollouts: $\lambda_b=10.0$, $[.27, .21, .25, .25]$; $\lambda_b=.01$, $[ 0., 0., .62, .38]$.}
    \label{fig:optionsovertime}
\end{figure}

\subsubsection{Convergence of Mixtures to Options   }   To show that our formulation of Mixture-of-Experts decomposes to options in the optimal case, we investigate the distributions of our policy-over-options. We find that across 40 trials, 100\% of activations fell within a reasonable error bound of deterministic selection across 1M samples. That is, in 40 total trials across 4 environments (Hopper-v1, HalfCheetah-v1, Walker2d-v1, RoboschoolHumanoidFlagrun-v1), policies were trained for 500 iterations (or 5k iterations in the case of RoboschoolHumanoidFlagrun-v1). We collected 25k samples at the end of each trial. Among the gating activations across the samples, we recorded the number of gating activations within the range $\{0 + \epsilon, 1-\epsilon \}$ for $\epsilon=0.1$. 100\% fell within this range. 98.72\% fell within range $\epsilon=1^{-3}$. Thus at convergence, both intuitively and empirically we can refer to our gating function over experts as the policy-over-options and each of the experts as options.

\subsubsection{Effect of Uniform Distribution Regularizer   }   We find that forcing a uniform distribution over options can potentially be harmful. This can be seen in the experiment in Figure~\ref{fig:optionsovertime}, where we evaluate the 4 option case with $\lambda_b=\{0.1,10\}$. However, relaxing the uniform constraint results in rapid performance increases, particularly in the HalfCheetah-v1 where we see increases in learning speed with 4 options.

There is an intuitive explanation for this. In the 4-option case, with a relaxed uniform distribution penalty, we allow options to drop out during training. In the case of Hopper and Walker tasks, generally 2 options drop out slowly over time, but in HalfCheetah, only one option drops out in the first 20 iterations with a uniform distribution remaining across the remaining options as seen in Figure~\ref{fig:gravtransfer}. We posit that in the case of HalfCheetah there is enough mutually exclusive information in the environment state space to divide across 3 options, quickly causing a rapid gain in performance, while the Hopper tasks do not settle as quickly and thus do not see that large gain in performance.

\subsubsection{Latent Structure in Expert Demonstrations   }   
\begin{figure}
\includegraphics[width=.47\textwidth]{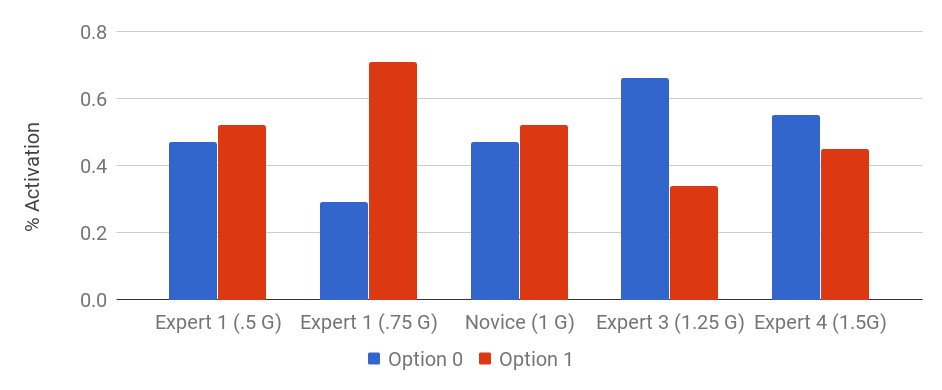}
\caption{Probability distribution of $\pi_\Omega$ over options on expert demonstrations. Inherent structure is found in the underlying demonstrations. The .75G demonstration state spaces are significantly assigned to Option 1 and similarly, the 1.25G state spaces to Option 0.}
\label{fig:expertsexplained}
\end{figure}

Another benefit of using options in the IRL transfer setting is that the underlying latent division of the original expert environments is learned by the policy-over-options. As seen in Figure~\ref{fig:expertsexplained}, the expert demonstrations have a clear separation among options. We suspect that options further away from the target gravity are not as specialized due to the fact that their state spaces are covered significantly by a mixture of the closer options (see supplemental material for supporting projected state space mappings). This indicates that the policy-over-options specializes over the experts and is thus inherently beneficial for use in one-shot transfer learning.

\section{Discussion}
We propose a direct extension of the options framework by adding joint reward-policy options. We learn these options in the context of generative adversarial inverse reinforcement learning and show that this method outperforms the single policy case in a variety of tasks -- particularly in transfer settings. Furthermore, the learned options demonstrate temporal and interpretable cohesion without specifying a call-and-return termination function.

Our formulation of joint reward-policy options as a Mixture Of Experts allows for: potential upscaling to extremely large networks as in \cite{superlargenets-17}, reward shaping in forward RL, and using similarly specialized MoEs in generative adversarial networks.
This work presents an effective and extendable framework. Our optionated networks capture the problem structure effectively, which allows strong generalization in one-shot transfer learning. Moreover, as adversarial methods are now commonly used across a myriad of communities, we believe the embedding of options within this methodology is an excellent delivery mechanism to exploit the benefits of hierarchical RL in many new fields.

\section{Acknowledgements}
We thank CIFAR, NSERC, The Open Philanthropy Project, and the AWS Cloud Credits for Research Program for their generous contributions.

\bibliographystyle{aaai}
\bibliography{biblio}

\onecolumn
\section{Supplemental Material}
\label{sec:appendix}
\input{appendix}

\end{document}

%% file: appendix.tex
\pdfoutput=1

\section{Expanded Equations}

The expectation over the discriminator loss for the option case can be expanded:

\begin{equation}
\begin{split}
L_\Omega &= \mathbb{E}_\omega \left[ \pi_{\Omega,\zeta}(\omega|s) L_{{\hat{\theta}},\omega} \right] + L_{reg}\\
&= \sum_\omega \pi_{\Omega,\zeta}(\omega|s) L_{{\hat{\theta}},\omega} + L_{reg}
\end{split}
\label{eq:discriminator_loss2}
\end{equation}

As can the regularization terms:

\begin{equation}
\begin{split}
    L_b &= \sum_\omega || \mathbb{E}[\pi_\Omega(\omega)] - \tau ||_2\\
    & \approx \sum_\omega || \frac{1}{m_b} \sum_i^{m_b} (\pi_\Omega(\omega| s_i)) - \tau||_2
\end{split}
\end{equation}

\begin{equation}
\begin{split}
    L_e &= \mathbb{E}\left[ ||  \left(\frac{1}{||\Omega||} \sum_\omega \pi_\Omega(\omega) \right) - \tau ||_2 \right]\\
    & \approx \frac{1}{m_b} \sum_i^{m_b}  || \left(\frac{1}{||\Omega||} \sum_\omega \pi_\Omega(\omega| s_i) \right) - \tau||_2
\end{split}
\end{equation}

\begin{equation}
\begin{split}
    L_v &= - \sum_\omega var_\omega\{\pi_\Omega(s)\}\\
    &\approx -\sum_\omega \frac{1}{m_b} \sum_i^{m_b} \left( \pi_\Omega (\omega| s_{i}) - \left( \frac{1}{m_b} \sum_i^{m_b} \pi_\Omega(\omega| s_{i})\right)\right)
    \end{split}
\end{equation}

\section{Expert Collection}

The expert demonstration rollouts (state sequences) for all OpenAI Gym \cite{gym} environments were obtained from policies trained for 1000 iterations using Trust Region Policy Optimization \cite{TRPO} with parameters $KL_{\max} = .01$, generalized advantage estimation $\lambda=.97$, discount factor $\gamma=.99$ and batch size $25,000$ (rollout timesteps shared when updating the discriminator and policy). For the Roboschool \cite{PPO} Flagrun-v1 environment, the rollouts were obtained using the PPO pre-trained expert provided with Roboschool.

\section{Experimental Setup and Hyperparameters}
Observations are \textbf{not} normalized in all cases as we found that it did not help or hurt performance. In all cases for advantage estimation we use a value approximator as in~\cite{TRPO}, which uses L-BFGS optimization with a mixing fraction of $.1$. That is, it uses the current prediction $V_\theta(s)$ and mixes it with the actual discounted returns with $.1$ belonging to the actual discounted returns and $.9$ belonging to the current prediction. This is identical to the original Trust Region Policy Optimization (TRPO) code as provided at: \url{https://github.com/joschu/modular_rl/}. We perform a maximum of 20 L-BFGS iterations per value function update. For all the environments, we let the agent act until the maximum allowed timesteps of the specific environment (as set by default in OpenAI Gym), gather the rollouts and keep the number of timesteps per batch desired. For all policy optimization steps in both IRLGAN and OPTIONGAN, we use TRPO with the with parameters set to the same values as the ones used for the expert collection ($KL_{\max} = .01$, generalized advantage estimation $\lambda=.97$, discount factor $\gamma=.99$ and batch size $25,000$), except for the Roboschool Flagrun Experiment where PPO was used instead, as explained in its respective section below.

\subsection{Simple Tasks and Transfer Tasks}

For simple tasks we use 10 expert rollouts while for transfer tasks we use 40 expert rollouts (10 from each environment variation).

\begin{figure*}[!htb]
\includegraphics[width=.33\textwidth]{"images/Hopper-v1__One_Shot_Transfer_"}
\includegraphics[width=.33\textwidth]{"images/HalfCheetah-v1__One_Shot_Transfer_"}
\includegraphics[width=.33\textwidth]{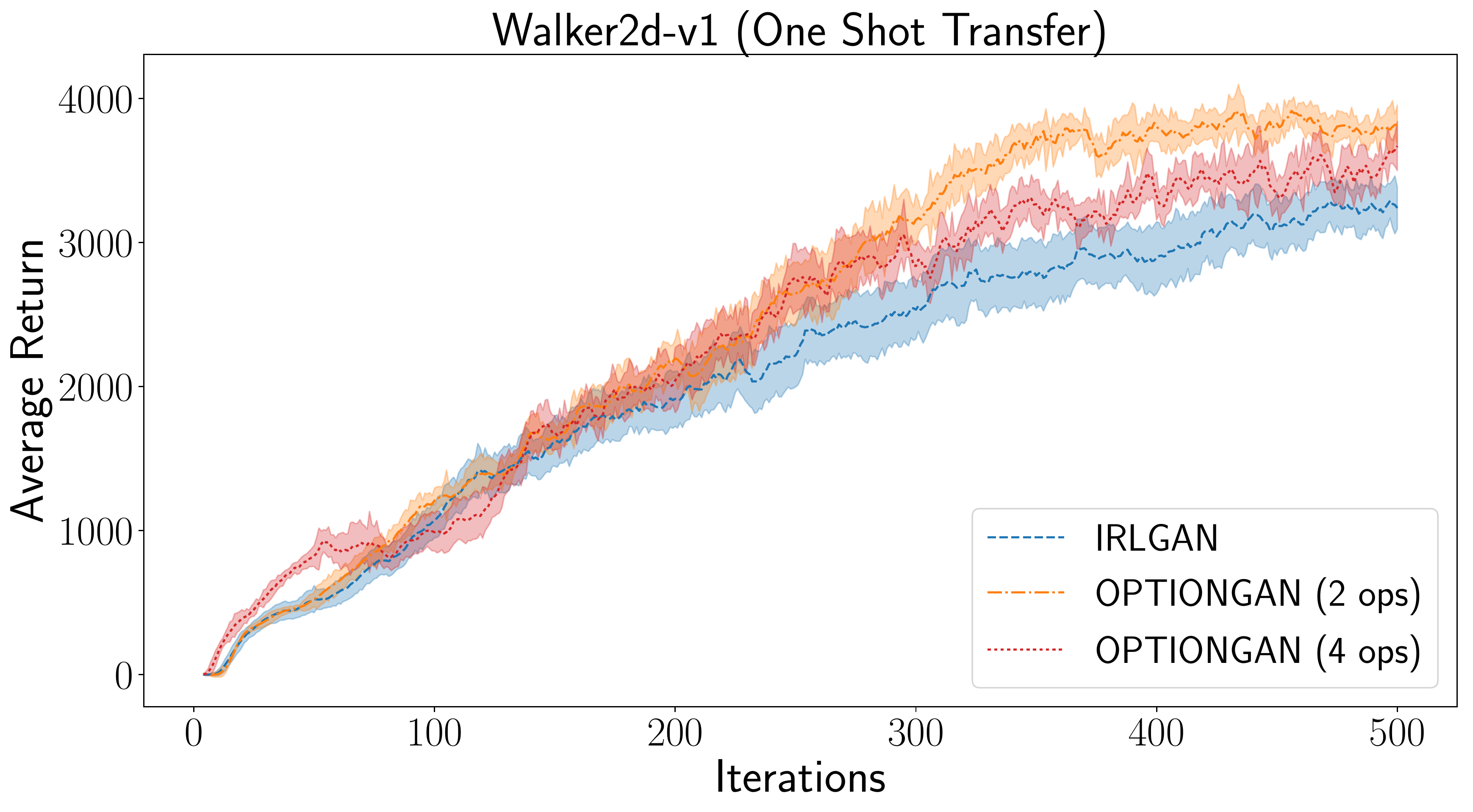}
\caption{Experiments from 40 expert demonstrations from varied gravity environments without any demonstrations on the novice training environment. Average returns from expert demonstrations across all mixed environments: 3657.71 (Hopper-v1), 4181.97 (HalfCheetah-v1), 4218.12 (Walker2d-v1).}
\label{fig:gravtransfer2}
\end{figure*}

\begin{figure*}[!htb]
\includegraphics[width=.33\textwidth]{images/Hopper-v1.pdf}
\includegraphics[width=.33\textwidth]{images/HalfCheetah-v1.pdf}
\includegraphics[width=.33\textwidth]{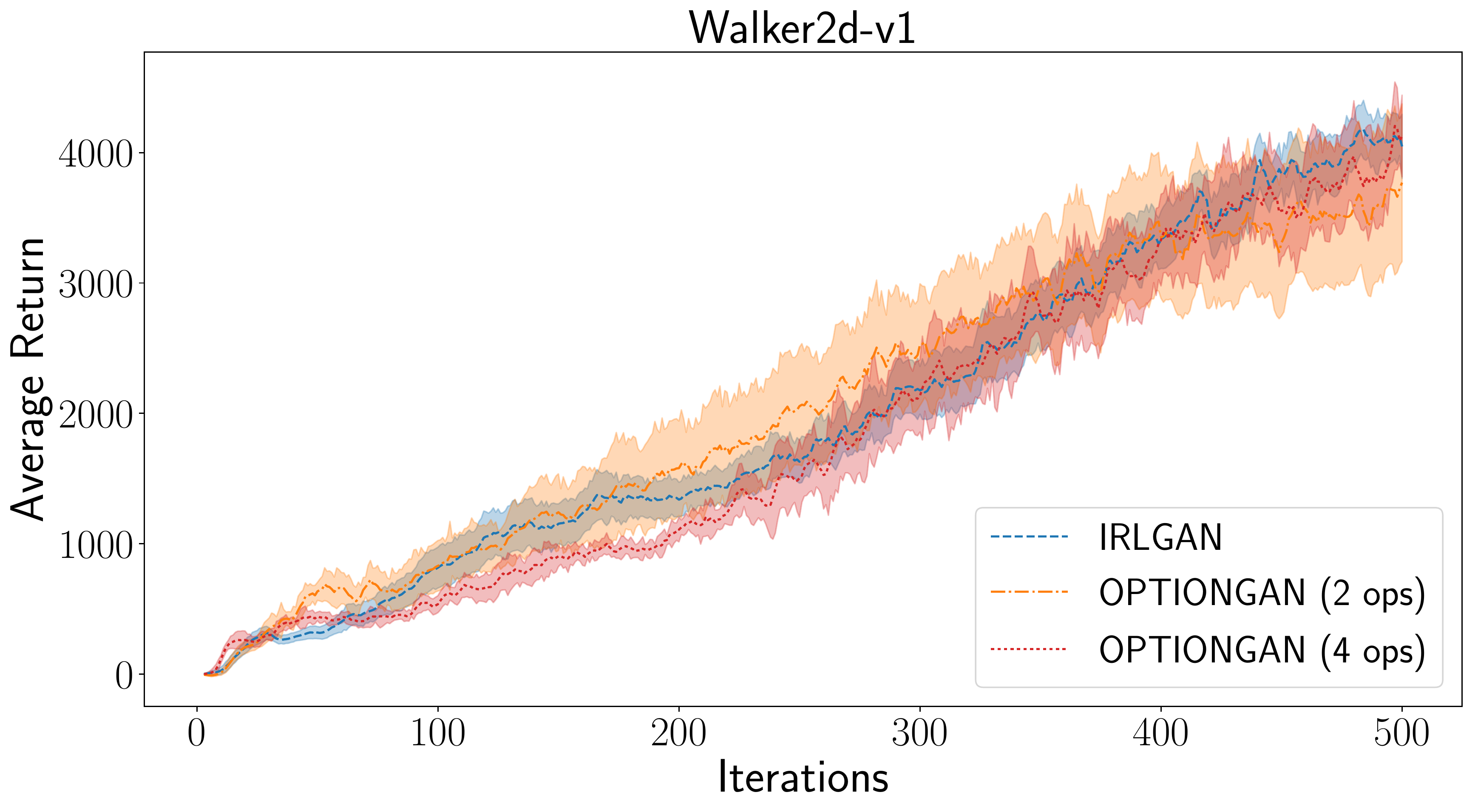}
\caption{Simple locomotion task learning curves. The True Average Return provided by the expert demonstrations are: 3778.82 (Hopper-v1), 4156.94 (HalfCheetah-v1), 5528.51 (Walker2d-v1). Error bars indicate standard error of True Average Return across 10 trial runs.}
\label{fig:simple2}.
\end{figure*}

\subsubsection{IRLGAN}
\begin{figure*}[!htb]
\includegraphics[width=.48\textwidth]{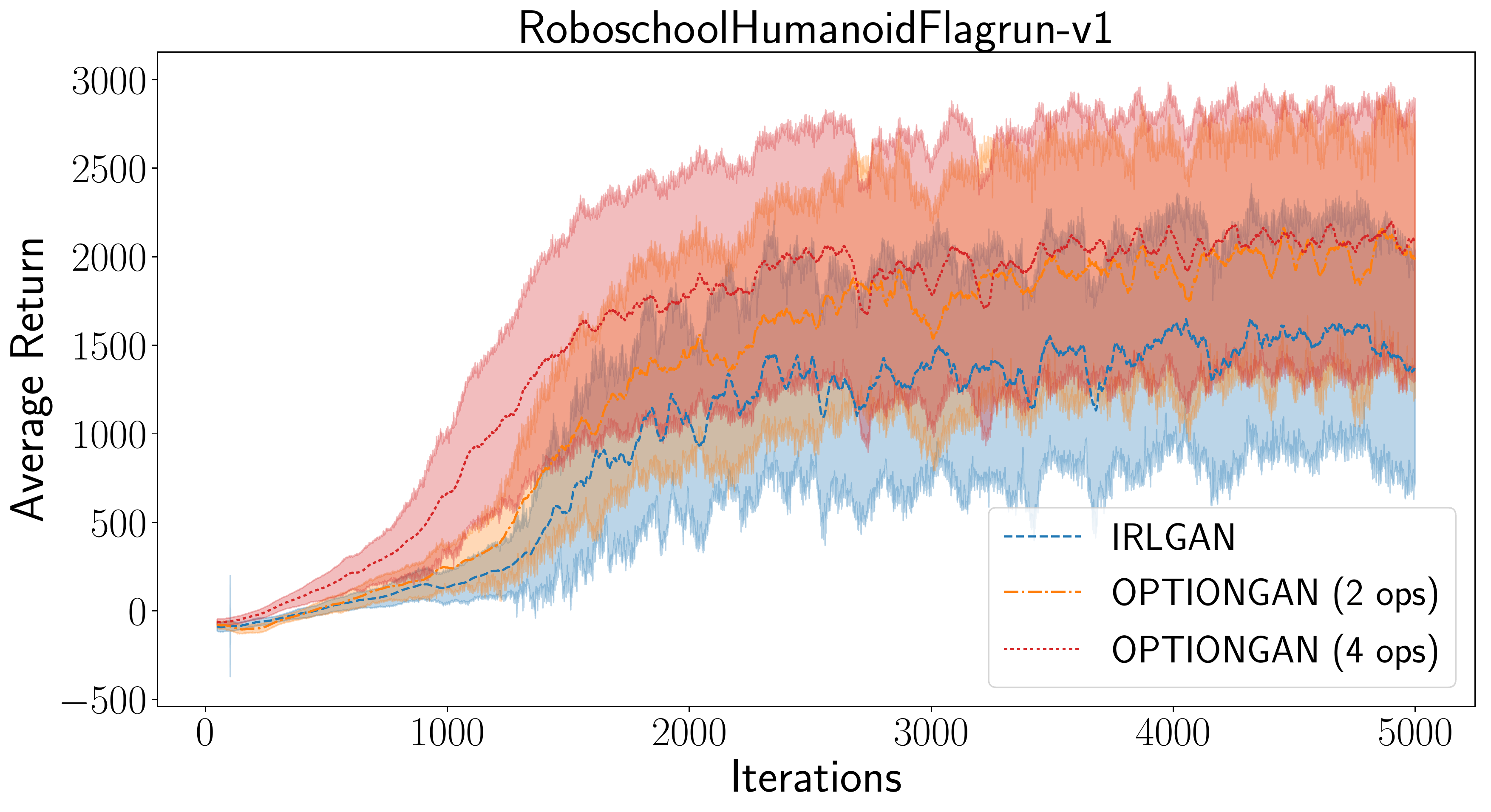}
\includegraphics[width=.48\textwidth]{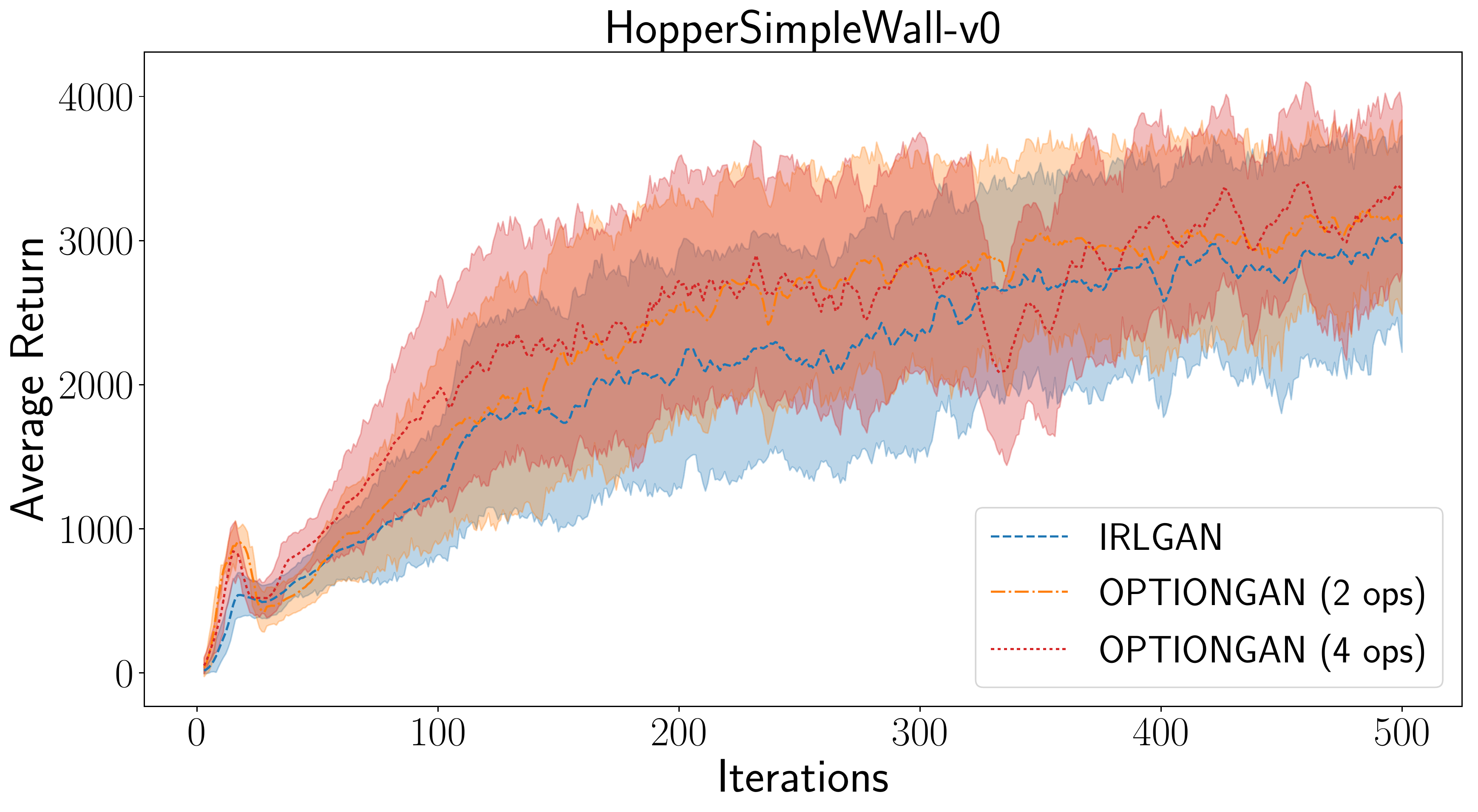}
\caption{Evaluation on RoboschoolHumanoidFlagrun-v1 environment~\cite{PPO}. Average expert rollout returns: 2822.13. }
\end{figure*}

We use a Gaussian Multilayer Perceptron policy as in~\cite{TRPO} with two 64 units hidden layers and tanh hidden layer activations. The output of the network gives the Gaussian mean and the standard deviation is modeled by a single learned variable as in~\cite{TRPO}. Similarly for our discriminator network, we use the same architecture with a sigmoid output, tanh hidden layer activations, and a learning rate of $1\cdot 10^{-3}$ for the discriminator. We do not use entropy regularization or $l2$ regularization as it resulted in worse performance. For every policy update we perform 3 discriminator updates as we found the policy optimization step is able to handle this and results in faster learning.

\subsubsection{OPTIONGAN}

Aligning with the IRLGAN networks, we make use of a Gaussian Multilayer Perceptron policy as in~\cite{TRPO} with 2 hidden layers of 64 units with tanh hidden layer activations for our shared hidden layers. These hidden layers connect to $||\Omega||$ options depending on the experiment (2 or 4). In this case the output of the network gives the Gaussian mean for each option and the standard deviation is modeled by a single learned variable per option. The policy-over-options is also modeled by a 2 layered 64 units network with tanh activations and a softmax output corresponding to the number of options. For our discriminator, we use the same architecture with tanh hidden layer activations, a sigmoid output  and $||\Omega||$ outputs, one for each option. We use the policy over options to create a specialized mixture of experts model with a specialized loss function which converges to options. We use a learning rate of $1\cdot 10^{-3}$ for the discriminator. Same as in IRLGAN, we do not make use of entropy regularization or $l2$ regularization as we found either regularizers to hurt performance. Instead we use scaling factors for the regularization terms included in the loss: $\lambda_b=10.0, \lambda_e =10.0, \lambda_v=1.0$ for the 2 options case and  $\lambda_b=0.01, \lambda_e =10.0, \lambda_v=1.0$ for the 4 options case. Again, we perform 3 discriminator updates per policy update

\subsection{RoboschoolHumanoidFlagrun-v1} \label{roboschool_details}

For Roboschool experiments we use proximal policy optimization (PPO) with a clipping objective~\cite{PPO} (clipping parameter set to $\epsilon = .02$). We perform 5 Adam~\cite{adam-kingma} policy updates on the PPO clipping objective with a learning rate of $1\cdot10^{-3}$. The value function and advantage estimation parameters from previous experiments are maintained while our network architecture sizes are increased to $(128,128)$ and use ReLU activations instead of tanh.

\section{Decomposition of Rewards over Expert Demonstrations}

We show that the trained policy-over-options network shows some intrinsic structure over the expert demonstrations.

\begin{figure}[h]
\centering\includegraphics[width=.51\textwidth]{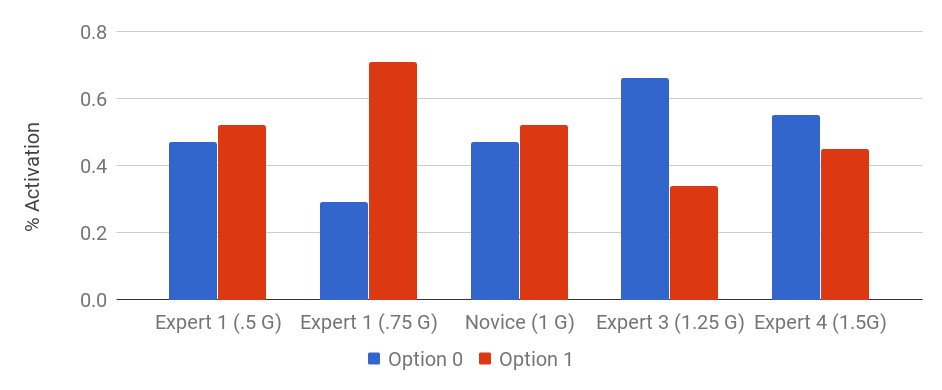}
\caption{Probability distribution of $\pi_\Omega$ over options on expert demonstrations. Inherent structure is found in the demonstrations.}

\label{fig:expertsexplained2}
\end{figure}

\begin{figure}[!htb]

\centering
\includegraphics[width=.3\textwidth]{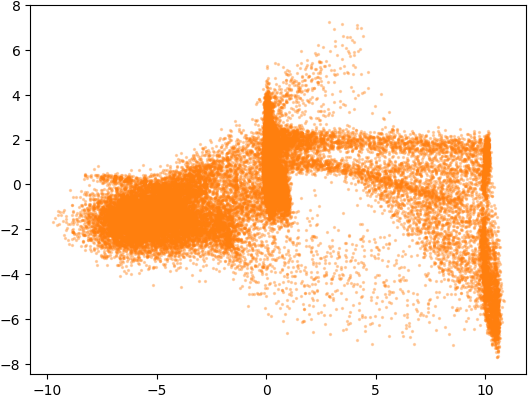}
\includegraphics[width=.3\textwidth]{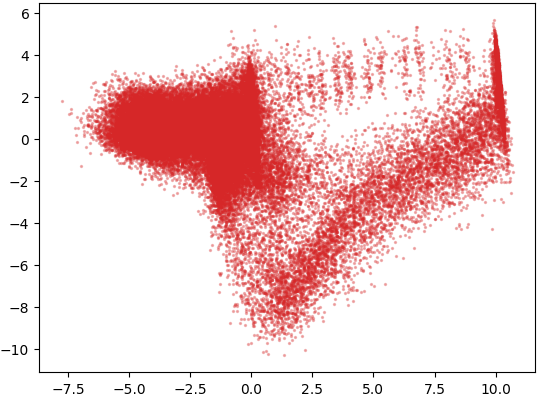}
\includegraphics[width=.3\textwidth]{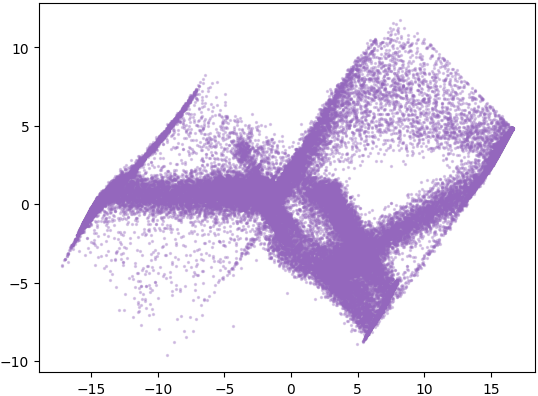}
\includegraphics[width=.3\textwidth]{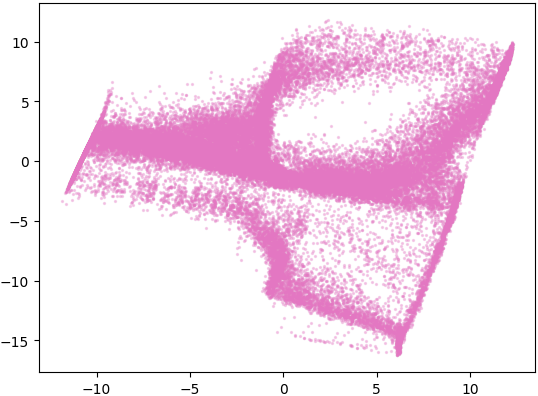}
\includegraphics[width=.3\textwidth]{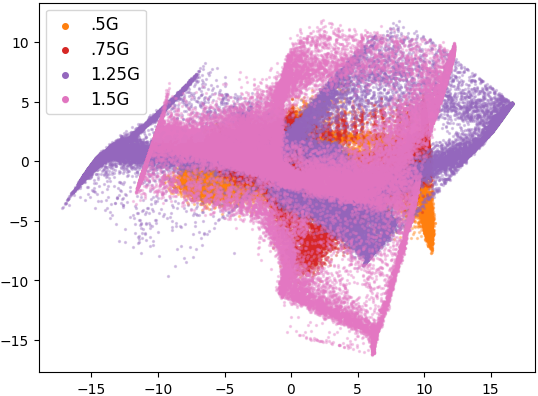}
\caption{State distribution of expert demonstrations projected onto a 2D plane to pair with Figure~\ref{fig:expertsexplained}. Axes correspond to projected state dimensions from the original 6-D state space using SVD.}

\label{fig:states2}
\end{figure}

In Figure~\ref{fig:expertsexplained2} are shown the activation of the gating function across expert rollouts after training. We see that the underlying division in expert demonstrations is learned by the policy-over-options, which indicates that our method for training the policy-over-options induces it to learn a latent structure to the expert demonstrations and thus can benefit in the transfer case since each option inherently specializes to be used in different environments. We find that options specialized more clearly over the experts with environments closest to the normal gravity environment, while the others use an even mixture of options. This is due to the fact that the mixing specialized options are able to cover the state space of the non-specialized options as we can observe from the state distribution of the expert demonstrations shown in Figure~\ref{fig:states2}.